\documentclass[10pt,twocolumn]{article}
\usepackage{redknotpaper}
\usepackage{amsmath}
\usepackage{amssymb}
\title{RedKnot-MLA: Multi-Head Offline--Online Reuse for\\DeepSeek-V4 Long-Context Serving}
\author{
 {\rm Yang Liu}\textsuperscript{$\ast$}$^\dagger$ \quad
 {\rm Zhaokai Luo}\textsuperscript{$\ast$}$^\dagger$ \quad
 {\rm Huayi Jin}$^\dagger$ \quad
 {\rm Ruozhou He}$^\dagger$ \quad
 {\rm Chenchen Hong}$^\dagger$ \quad
 {\rm Mingxiao Ma}$^\dagger$ \quad\\
 {\rm Biao Zhang}$^\dagger$ \quad
 {\rm Zhiyong Wang}$^\dagger$ \quad
 {\rm Boyu Wang}$^\mathparagraph$ \quad
 {\rm Guanjie Chen}$^\mathparagraph$ \quad
 {\rm Yifei Liu}$^\clubsuit$ \quad
 {\rm Tao Xie}$^{\mathsection\ddagger}$ \quad
 {\rm Junhao Hu}$^{\mathsection\ddagger}$\\
 $^\dagger${\it Xiaohongshu Inc., China} \quad
 $^\ddagger${\it Peking University} \quad
 $^\mathparagraph${\it Huawei Cloud} \quad
 $^\clubsuit${\it Shanghai Jiao Tong University}\\
 $^\mathsection${\it Beijing Tongming Lake Information Technology Application Innovation Center}\\[0.6ex]
 {\normalsize\rm
   \textsuperscript{$\ast$}Corresponding to:\;
   Yang Liu~\href{mailto:xiaoyi52@xiaohongshu.com}{\textless xiaoyi52@xiaohongshu.com\textgreater} or\;
   Zhaokai Luo~\href{mailto:luozhaokai@xiaohongshu.com}{\textless luozhaokai@xiaohongshu.com\textgreater}%
 }\\
\url{github: https://github.com/rednote-machine-learning/RedKnot}}
\date{}

\begin{document}
\maketitle
\raggedbottom

\begin{abstract}
Multi-head latent attention (MLA) exposes many logical query heads through one
packed latent KV stream.  This representation is memory efficient, but it
removes the physical per-head cache boundary assumed by conventional head-wise
reuse.  We present \system{}, a DeepSeek-V4 realization of RedKnot's
head-aware reuse principle.  Each immutable document is processed offline at
canonical position zero; certified Local-head contributions are retained as
\emph{MLA-Off}.  At serving time, query-side RoPE relocation restores the
document's request position, a small Global-head set and protected Local token
rows are recomputed as \emph{MLA-Online}, and the two paths are merged before a
single shared output projection.  The packed MLA latent is never split.
DeepSeek-V4-Flash uses 37 reusable layers and a 56/8 Local/Global partition,
giving a 75.29\% analytic logical head-row ceiling; the Pro-0813 profile uses
55 layers and 112/16 heads, giving 78.89\%.  Frozen Flash operating points show
hot-artifact TTFT speedups of $2.02$--$3.84\times$.  At 256K, the archived
three-dataset study reports an aggregate F1 change of $+3.24$ percentage points,
an EM change of $+4.16$ points, and a 78.7--79.5\% analytic major-operator
arithmetic saving, while one dataset decreases by 2.81 F1 points.  A separate
author-reported 256K hot-artifact QPS measurement is approximately $2.0\times$;
because its raw concurrency trace is not included in this bundle, we mark it
as preliminary rather than archived evidence.  We describe the factorization,
position repair, token-row closure, sparse-MoE support, TP8 integration, and the
measurement boundaries needed to interpret these results.
\end{abstract}

\section{Introduction}

Non-prefix KV reuse can avoid repeatedly encoding retrieved documents in RAG
and agentic workloads.  CacheBlend~\citep{yao2025cacheblend},
CacheSlide~\citep{liu2026cacheslide}, EPIC~\citep{hu2025epic}, and
HYPIC~\citep{liu2026hypic} repair or link cached segments after their context
changes.  RedKnot instead decomposes reuse at attention-head granularity:
document-stable heads can be reused while request-sensitive heads remain
online~\citep{redknot2026}.  Applying that idea to DeepSeek-V4 is not a
mechanical port.  MLA has many logical query heads but only one physical packed
latent KV stream~\citep{deepseekv2,deepseekv3,deepseek2026v4}; manufacturing a
physical cache per logical head would duplicate the latent state and defeat
the native representation.

\system{} preserves the packed latent and factorizes the logical-head
\emph{output contribution}.  Local heads are evaluated per document at
canonical position zero and stored as a projected offline artifact.  The
online request rotates queries into each document's canonical frame, evaluates
one Global head per native output group, repairs selected Local token rows, and
combines document segments with a stable segmented merge.  A group-aligned
sliced projection makes the Local and Global contributions additive before the
shared second output projection.  Indexer-guided token-row closure and a
conservative adaptive MoE policy reduce the token-wide work that would
otherwise remain after head reuse; they support, rather than replace, the
defining multi-head decomposition.

\begin{figure*}[t]
  \centering
  \includegraphics[width=0.99\textwidth]{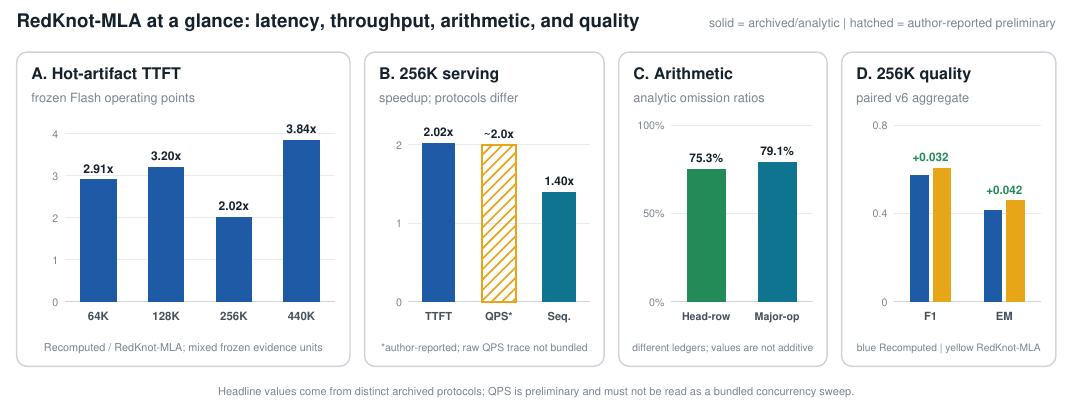}
  \caption{Performance overview.  Panel A shows frozen hot-artifact TTFT
  diagnostics.  Panel B separates the archived 256K TTFT ratio, the
  author-reported preliminary QPS ratio (hatched), and an archived sequential
  request-rate diagnostic; the latter used unequal output-token totals and is
  not a concurrency-QPS result.  Panel C reports two non-additive analytic
  omission ratios.  Panel D reports paired aggregate F1/EM at 256K.}
  \label{fig:overview}
\end{figure*}

Figure~\ref{fig:overview} gives the evidence-aware headline.  It deliberately
keeps four quantities separate.  TTFT is an archived hot-artifact diagnostic;
the 256K QPS bar is the author's approximately $2\times$ result but its raw
concurrency trace is not bundled; the compute bars are analytic ledgers rather
than board-level achieved FLOPs; and the quality bars come from the paired
three-dataset v6 study.  These values therefore describe an attainable
operating envelope, not one homogeneous scaling experiment.

The report makes four contributions:
\begin{itemize}
  \item an exact logical-head factorization that leaves MLA's physical packed
        latent KV representation unchanged;
  \item canonical-position offline artifacts with online RoPE relocation,
        Global-head recomputation, and protected-row correction;
  \item concrete Flash and Pro profiles plus fused prefetch, validation, and
        merge integration for an eight-GPU tensor-parallel runtime; and
  \item a protocol-qualified evaluation that distinguishes measured latency
        and quality, preliminary QPS, and analytic compute accounting.
\end{itemize}

\section{MLA-Off/MLA-Online Design}

\subsection{Logical heads, one physical latent}

Let $O\in\mathbb{R}^{T\times H\times d}$ be the logical attention-head output
of an MLA layer.  We partition the logical heads into disjoint Local and Global
sets,
\begin{equation}
  L\cap G=\emptyset,\qquad L\cup G=\{1,\ldots,H\}.
  \label{eq:partition}
\end{equation}
Local and Global are \system{} execution roles, not physical MLA KV-head
types.  The shared packed latent $C$ remains native and is paged exactly once.
The first output projection is bias free and can be sliced by the disjoint
logical-head channel ranges into $W_{a,L}$ and $W_{a,G}$.  With inverse output
RoPE denoted by $R^{-1}$, define
\begin{align}
  z_{\mathrm{off}} &= W_{a,L}R^{-1}(O_L), \\
  z_{\mathrm{online}} &= W_{a,G}R^{-1}(O_G).
  \label{eq:offonline}
\end{align}
The projected state and layer output are
\begin{equation}
  z=z_{\mathrm{off}}+z_{\mathrm{online}},\qquad y=W_bz.
  \label{eq:factorization}
\end{equation}
Equation~\ref{eq:factorization} is an exact linear identity when $O_L$ and
$O_G$ equal full-recomputation outputs.  Approximation enters only when a Local
state is restored in a changed request context.  Global heads and protected
Local rows provide the correction path.

\begin{figure*}[t]
  \centering
  \includegraphics[width=0.99\textwidth]{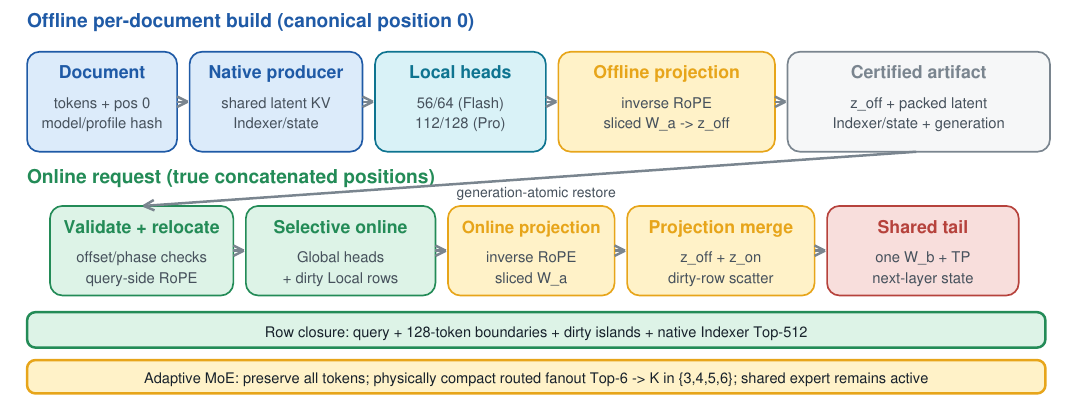}
  \caption{\system{} preserves DeepSeek-V4's single packed MLA latent stream.
  Offline, each independently positioned document produces a certified
  Local-head contribution.  Online, query-side RoPE relocation, Global-head
  recomputation, and protected-row replay produce the complementary state.
  Both paths meet in sliced $W_a$ space before one shared $W_b$ projection.}
  \label{fig:architecture}
\end{figure*}

\subsection{Canonical offline artifacts}

Every immutable document is tokenized and evaluated independently with its
first content token at position zero.  This makes an artifact independent of
the document's eventual order in a request.  In each reusable layer, the
producer retains the native packed latent, projected Local contribution,
Indexer/controller metadata, boundary metadata, and a certificate binding the
checkpoint, tokens, profile, tensor-parallel geometry, and artifact generation.
Artifacts are immutable and generation-atomic.

\begin{rkalgorithm}{Build one canonical MLA-Off document artifact}
\label{alg:offline}
\Require document tokens $x$, profile $\Pi$, checkpoint identity $h_m$
\Ensure certified immutable artifact $A$
\State $p\gets(0,1,\ldots,|x|-1)$
\State $(C,\{O_L^{(l)}\},I,B)\gets\Call{NativeProducer}{x,p,\Pi}$
\ForAll{reusable layers $l\in\mathcal{R}$}
  \State $z_{\mathrm{off}}^{(l)}\gets
    W_{a,L}^{(l)}R^{-1}(O_L^{(l)})$
\EndFor
\State $k\gets\Call{Hash}{h_m,x,\Pi,\mathrm{TP},\mathrm{generation}}$
\State $A\gets\Call{AtomicPublish}{k,C,z_{\mathrm{off}},I,B}$
\State \Return $A$
\end{rkalgorithm}

No online request is allowed to consume $A$ unless every TP rank validates the
same certificate.  Missing layers, token/profile mismatch, stale generation,
incompatible TP geometry, or rank disagreement causes an explicit fallback to
full recomputation before any reuse-specific collective is entered.

\subsection{Online position restoration and segmented merge}

Suppose document $i$, built at positions $0,\ldots,n_i-1$, appears at request
offset $\Delta_i$.  RoPE orthogonality gives
\begin{equation}
  \langle R(p_q)q,R(p_k+\Delta_i)k\rangle
  =\langle R(p_q-\Delta_i)q,R(p_k)k\rangle.
  \label{eq:rope}
\end{equation}
The runtime therefore rotates the online query by $-\Delta_i$, attends directly
to the canonical latent $C_i$, restores the output phase, and combines segment
statistics with a numerically stable log-sum-exp merge.  It need not rewrite or
physically concatenate each latent cache.

The Local contribution is not restored blindly.  The protected token-row set is
\begin{equation}
 \mathcal{A}=Q\cup N\cup D\cup B_{128}\cup I_{\mathrm{top}}
              \cup T_{32\mathrm{K}},
 \label{eq:protected}
\end{equation}
where $Q$ contains query rows, $N$ new rows, $D$ dirty rows, $B_{128}$ the first
and last 128 rows of each relocated segment, $I_{\mathrm{top}}$ rows selected by
the native Indexer recomputed in the current request, and $T_{32\mathrm{K}}$ the
recent 32K tail.  A fixed controller may expand the active budget to 10\%, 20\%,
or 25\% of candidate rows.  Cached Local values on $\mathcal{A}$ are
\emph{replaced}, never double counted.

\begin{rkalgorithm}{Relocate, recompute, and merge one online layer}
\label{alg:online}
\Require layer $l$, artifacts $\{A_i\}$, offsets $\{\Delta_i\}$, request $r$
\Ensure layer output $y^{(l)}$
\State $\Call{AllRanksValidate}{\{A_i\},l,r,\Pi}$
\State $\mathcal{A}\gets\Call{ProtectedRows}{r,\{A_i.I_l,A_i.B_l\}}$
\State $\Call{WaitPrefetch}{\mathrm{LayerGroup}(l)}$
\ForAll{document segments $i$}
  \State $q_i\gets R(-\Delta_i)q$
  \State $(O_{G,i},O_{L,\mathcal{A},i},s_i)\gets
    \Call{SelectiveMLA}{q_i,A_i.C_l,\mathcal{A}}$
  \State $(\widetilde O_{G,i},\widetilde O_{L,\mathcal{A},i})\gets
    \Call{RestorePhase}{O_{G,i},O_{L,\mathcal{A},i},\Delta_i}$
\EndFor
\State $(O_G,O_{L,\mathcal{A}})\gets
  \Call{SegmentedLSEMerge}{\{\widetilde O_{G,i},
  \widetilde O_{L,\mathcal{A},i},s_i\}}$
\State $z_{\mathrm{online}}\gets W_{a,G}^{(l)}R^{-1}(O_G)
  +W_{a,L}^{(l)}R^{-1}(O_{L,\mathcal{A}})$
\State $z\gets\Call{RestoreAndReplace}{\{A_i.z_{\mathrm{off},l}\},
  z_{\mathrm{online}},\mathcal{A}}$
\State \Return $W_b^{(l)}z$
\end{rkalgorithm}

\subsection{Adaptive sparse MoE as a supporting path}

Head reuse alone does not remove token-wide FFN work.  For active rows in
reusable middle layers, the native router first produces its Top-6 routed
experts with weights $p_{(1)}\ge\cdots\ge p_{(6)}$.  We normalize these six
weights only to decide the retained fanout,
\begin{align}
 \widehat p_j&=\frac{p_{(j)}}{\sum_{r=1}^{6}p_{(r)}},\\
 K_t&=\min\left\{k\in\{3,4,5,6\}:\right.\nonumber\\[-2pt]
 &\hspace{34mm}\left.\sum_{j=1}^{k}\widehat p_j\ge\tau\right\}.
 \label{eq:moe}
\end{align}
The executed experts keep their original router weights $p_{(j)}$; they are not
renormalized.  The shared expert is always retained, and all full-recompute
fence layers keep native Top-6 routing.  Thus an ambiguous router distribution
automatically uses more experts, while a concentrated distribution can use as
few as three.  Head-row, token-row, and expert-fanout savings overlap and are
never added as independent percentages.

\section{Profiles and Runtime Integration}

\subsection{DeepSeek-V4 Flash and Pro profiles}

Both profiles keep the first and last three transformer layers as
full-recompute fences.  In the middle region, the Local/Global split aligns
with DeepSeek-V4's native output groups: every group of eight logical heads
keeps one Global head online and assigns seven Local heads to MLA-Off.  Flash
therefore uses 3 full-recompute, 37 reusable, and 3 full-recompute layers
(3F--37R--3F), with 56/8 Local/Global heads.  Pro-0813 uses 3F--55R--3F and
112/16 heads.  The structural logical head-row ceiling is
\begin{equation}
 S_{\mathrm{head}}^{\max}
 =\frac{N_R}{N}\frac{|L|}{H},
 \label{eq:ceiling}
\end{equation}
which is 75.29\% for Flash and 78.89\% for Pro.  These are eligibility ceilings,
not whole-model FLOP or time savings.  Flash has paired long-context evidence;
Pro currently has TP8 load/HTTP smoke validation only.

\begin{figure*}[t]
  \centering
  \includegraphics[width=0.99\textwidth]{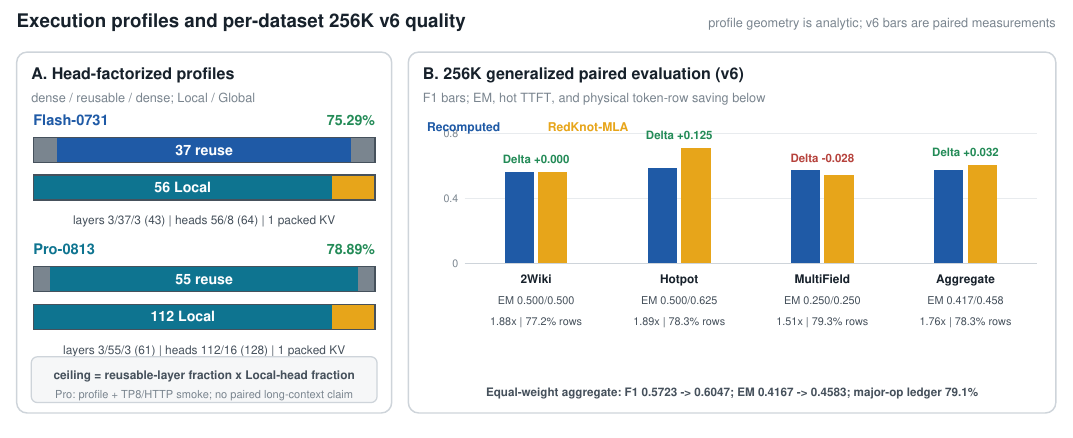}
  \caption{Profile geometry and the paired 256K v6 dataset breakdown.  The
  left panel is analytic and preserves one physical packed KV stream.  The
  right panel is measured by dataset: bars show F1, annotations show F1 change,
  EM, hot-artifact TTFT speedup, and physical token-row saving.  Pro geometry
  must not be read as a Pro long-context performance result.}
  \label{fig:profile-quality}
\end{figure*}

\subsection{Turning omitted work into latency reduction}

The TP8 fast path groups restoration and online execution rather than treating
cache management as a sequence of small kernels.  Artifacts are stored by
layer group, TP rank, and projected output-channel range.  While layer $l$ is
running, a dedicated CUDA stream prefetches tiles for a future layer group;
device events, not host polling, guard consumption.  Repeated document bundles
reuse compact segment-pointer geometry.

The consumer fuses query-side RoPE views, Global-head and protected-row
attention, sliced $W_a$ projection, Local restoration, and row-wise replacement
into a contiguous projected state.  The shared $W_b$ projection runs once.
This avoids repeated HBM materialization, scatter/gather kernels, and extra TP
barriers that can otherwise erase arithmetic savings.  SegPagedAttention
provides the segment placement abstraction while the underlying MLA latent
remains packed; it does not create physical pages per logical head.

The runtime also records four non-interchangeable namespaces: logical
head-row saving, physical token-row saving, analytic major-operator arithmetic
saving, and measured wall-clock timing.  Certification counters additionally
make any fallback visible.  A result is rejected if the RedKnot-MLA arm silently
runs full recomputation.

\section{Evaluation}

\subsection{Setup and protocol}

Experiments use one local DeepSeek-V4-Flash checkpoint on eight NVIDIA H200
GPUs with TP=8 and DP=CP=PP=1.  The archived environment uses CPython 3.11.13,
PyTorch 2.9.1 with CUDA 12.8, FlashMLA 1.0.0, and SGL Kernel 0.3.20.  TTFT is
measured from model-forward entry to first-token availability after runtime and
kernel warm-up.  Host snapshot preparation and offline artifact construction
are excluded from hot-artifact TTFT for both arms.

At each length, the release suite contains 15 frozen records: ten short-answer
cases for task-quality aggregation and five longer-output cases for paired text
inspection.  The Recomputed arm processes the complete request online without
a hidden prefix advantage.  The RedKnot-MLA arm treats document one as a prefix;
the remaining documents were independently built at position zero and are
relocated online.  Both arms share exact tokens, offsets, checkpoint, generation
parameters, output cap, TP geometry, and timing boundary.  The four headline
TTFT points are independently frozen diagnostics rather than samples from one
unbiased scaling sweep.

\subsection{Hot-artifact performance and modeled boundaries}

\begin{figure*}[t]
  \centering
  \includegraphics[width=0.99\textwidth]{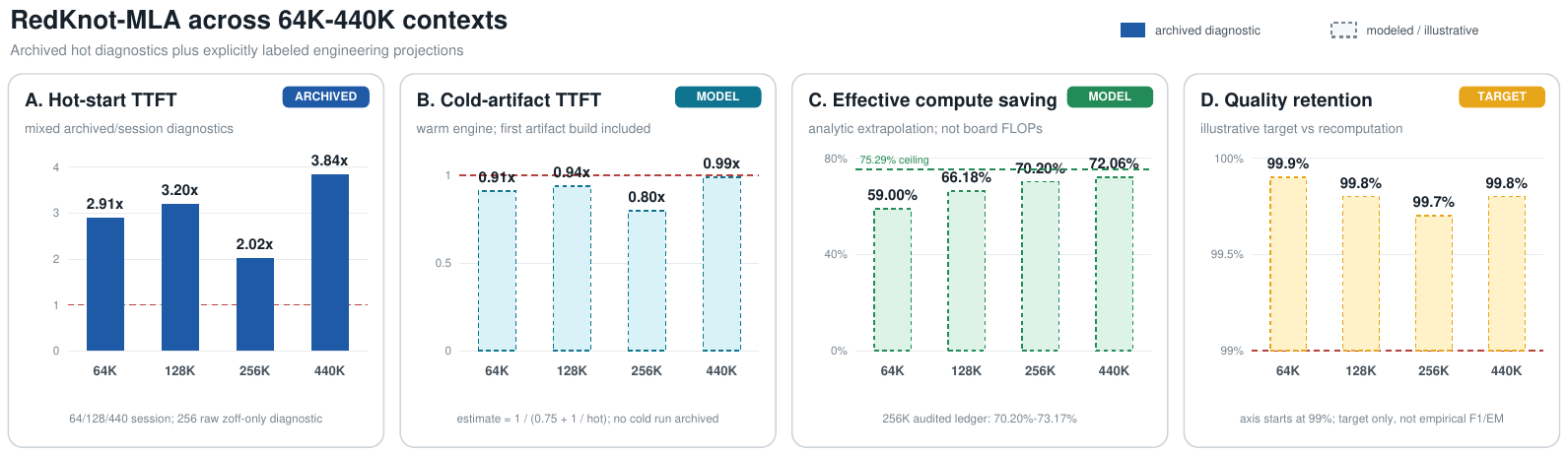}
  \caption{Evidence-aware 64K--440K engineering envelope.  Only panel A is an
  archived hot-artifact diagnostic.  Panel B is a cold-artifact model with
  $S_{\mathrm{cold}}=1/(0.75+1/S_{\mathrm{hot}})$ and no archived cold run.
  Panel C is an analytic extrapolation anchored by the 256K 70.20--73.17\%
  ledger, not board FLOPs.  Panel D is a 99\%--100\% design target, not
  empirical F1/EM.}
  \label{fig:envelope}
\end{figure*}

Figure~\ref{fig:envelope} preserves the user's four-length engineering view but
labels its evidence class on every panel.  The hot-artifact points are archived
diagnostics: $2.91\times$, $3.20\times$, $2.02\times$, and $3.84\times$ at
64K, 128K, 256K, and 440K.  Their non-monotonicity is expected because the
evidence units and fixed overheads differ; the figure must not be interpreted
as a controlled scaling law.

Cold-artifact latency includes first-time artifact construction and therefore
does not inherit the hot result.  The simple planning model in panel B assumes
construction contributes 0.75 normalized recomputation units, yielding
$0.80$--$0.99\times$; these values are projections, not measurements.  Likewise,
the 59.00\%, 66.18\%, 70.20\%, and 72.06\% curve in panel C models the growing
share of reusable work with context length.  The underlying Flash structural
ceiling remains length independent at 75.29\%.  The 256K first-document-prefix
sensitivity ledger independently reports 70.20--73.17\% as the active-row
allowance moves from 25\% to 10\%.

The author additionally reports approximately $2.0\times$ hot-artifact QPS at
256K.  We retain that point in Figure~\ref{fig:overview} because throughput is
an important serving outcome, but render it as preliminary.  The current bundle
does not contain its concurrency, arrival/batching policy, fixed generation
length, completed-request counts, or raw timing window.  The archived sequential
service-rate diagnostic is $0.031015\rightarrow0.043408$ requests/s
($1.40\times$), but the arms produced 110 versus 84 output tokens, so it is not
a fair QPS substitute.  A publication-grade QPS claim requires fixed-output
workloads at several concurrency levels and confidence intervals.

\subsection{256K generalization and quality}

Figure~\ref{fig:profile-quality} closes the cross-dataset loop.  For 2WikiMQA,
HotpotQA, and MultiFieldQA-en, F1 changes are respectively $0.0000$, $+0.1250$,
and $-0.0281$; EM changes are $0$, $+0.125$, and $0$.  The equal-weight aggregate
changes from 0.5723 to 0.6047 F1 and from 0.4167 to 0.4583 EM.  Mean hot-artifact
TTFT speedup is $1.76\times$, mean physical token-row saving is 78.3\%, and the
separately calculated major-operator ledger is 78.7--79.5\% (79.1\% mean).
The aggregate improves, but the 2.81-point MultiFieldQA-en F1 decrease prevents
a universal ``less than one point'' quality claim.

The supplementary figure isolates the sparse-MoE support path rather than
crediting its result to full RedKnot-MLA.  Across 50 8K examples, physical K3
changes aggregate F1 from 0.5905 to 0.5793 with unchanged EM of 0.42; its paired
bootstrap 95\% interval is $[-0.0800,+0.0569]$.  On five paired 256K prompts,
MoE-only model prefill improves from 14.918 s to 13.885 s ($1.074\times$).
This modest isolated gain explains why system-level head and token-row omission,
not expert pruning alone, drives the long-context TTFT result.

\begin{figure*}[t]
  \centering
  \includegraphics[width=0.99\textwidth]{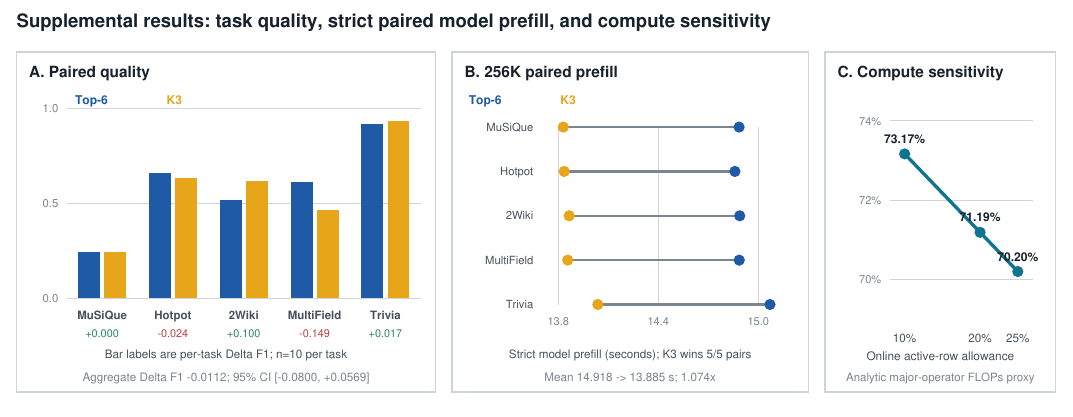}
  \caption{Supporting sparse-execution evidence.  Panels A and B are MoE-only
  studies and must not be presented as end-to-end RedKnot-MLA performance.
  Panel C is an analytic first-document-prefix compute sensitivity curve.}
  \label{fig:supplemental}
\end{figure*}

\balance
\section{Reproducibility and Limitations}

The public repository packages frozen JSONL suites, profile files, sparse-FFN
parameters, the benchmark driver, environment manifest, and a one-command
wrapper:
\begin{small}
\begin{verbatim}
git clone \
  git@github.com:rednote-machine-learning/RedKnot.git
cd RedKnot/test/srt/redknot
./run_deepseek_v4_flash_reproduction.sh
\end{verbatim}
\end{small}
The wrapper executes the 64K, 128K, 256K, and 440K suites, prints Recomputed and
RedKnot-MLA outputs side by side, and reports TTFT plus analytic compute ledgers.
It validates the environment and data hashes, uses a release lock to prevent two
TP8 owners, stops only its recognized GPU holder, preserves failure logs, and
restores the holder when the suite exits.  Each record binds the checkpoint and
profile hashes, token hash, segment offsets, artifact generation, active-row
policy, expert fanout, warm-up/repeat counts, timing scope, fallback counters,
and generated text.

Three limitations define the present claim boundary.  First, the four length
points are curated diagnostics with mixed evidence units; larger randomized
per-dataset samples and AB/BA repetitions are needed for population estimates.
Second, the QPS point is author reported but not yet trace backed in this bundle;
multi-concurrency, fixed-output sweeps must replace it before it is called a
reproducible measurement.  Third, compute values are analytic ledgers over named
operators.  Hardware counters are required to report achieved FLOPs, HBM traffic,
collective time, energy, and utilization.  Pro-0813 likewise needs paired
long-context quality and latency runs before its 78.89\% structural ceiling can
be connected to serving performance.

\section{Conclusion}

\system{} adapts head-aware reuse to DeepSeek-V4 MLA without fragmenting the
packed latent KV representation.  Independently built Local-head artifacts are
relocated with query-side RoPE, combined with Global-head and protected-row
recomputation, and merged before one shared projection.  Flash uses a
3F--37R--3F, 56/8 Local/Global profile; Pro uses 3F--55R--3F and 112/16.  The
archived Flash points demonstrate multi-fold hot-artifact TTFT reduction, while
the 256K paired study shows large analytic arithmetic omission with aggregate
F1/EM preservation and a visible per-dataset regression.  The preliminary
approximately $2\times$ QPS point is promising, but its raw concurrency trace
remains an explicit release task.  Keeping those evidence boundaries visible
turns the report into a reproducible systems description rather than a single
unqualified headline.

\bibliographystyle{abbrvnat}
\bibliography{references}

@inproceedings{yao2025cacheblend,
  author    = {Jiayi Yao and Hanchen Li and Yuhan Liu and
               Siddhant Ray and Yihua Cheng and Qizheng Zhang and
               Kuntai Du and Shan Lu and Junchen Jiang},
  title     = {{CacheBlend}: Fast Large Language Model Serving for
               {RAG} with Cached Knowledge Fusion},
  booktitle = {Proceedings of the Twentieth European Conference
               on Computer Systems (EuroSys '25)},
  year      = {2025},
  pages     = {94--109},
  publisher = {ACM},
  doi       = {10.1145/3689031.3696098}
}

@inproceedings{liu2026cacheslide,
  author    = {Yang Liu and Yunfei Gu and Liqiang Zhang and
               Chentao Wu and Guangtao Xue and Jie Li and
               Minyi Guo and Junhao Hu and Jie Meng},
  title     = {{CacheSlide}: Unlocking Cross Position-Aware
               {KV} Cache Reuse for Accelerating {LLM} Serving},
  booktitle = {24th USENIX Conference on File and Storage
               Technologies (FAST '26)},
  year      = {2026},
  pages     = {83--99},
  publisher = {USENIX Association},
  month     = feb
}

@inproceedings{hu2025epic,
  author    = {Junhao Hu and Wenrui Huang and Weidong Wang and
               Haoyi Wang and Tiancheng Hu and Zhang Qin and
               Hao Feng and Xusheng Chen and Yizhou Shan and Tao Xie},
  title     = {{EPIC}: Efficient Position-Independent Caching for
               Serving Large Language Models},
  booktitle = {Proceedings of the 42nd International Conference
               on Machine Learning},
  volume    = {267},
  pages     = {24391--24402},
  year      = {2025},
  publisher = {PMLR}
}

@article{liu2026hypic,
  author  = {Yifei Liu and Juntong Wu and Yang Liu and Junhao Hu and
             Minghao Li and Xiaoxu Chen and Weihang Chen},
  title   = {{HYPIC}: Accelerating Hybrid-Attention {LLM} Serving with
             Position-Independent Caching},
  journal = {arXiv preprint arXiv:2607.01299},
  year    = {2026}
}

@techreport{deepseek2026v4,
  author      = {{DeepSeek-AI}},
  title       = {{DeepSeek-V4}: Towards Highly Efficient
                 Million-Token Context Intelligence},
  institution = {DeepSeek-AI},
  year        = {2026},
  note        = {Technical Report},
  url         = {https://huggingface.co/deepseek-ai/DeepSeek-V4-Pro/blob/main/DeepSeek_V4.pdf}
}

@article{deepseekv2,
  title={{DeepSeek-V2}: A Strong, Economical, and Efficient Mixture-of-Experts Language Model},
  author={{DeepSeek-AI}},
  journal={arXiv preprint arXiv:2405.04434},
  year={2024}
}

@misc{redknot2026,
  title={{RedKnot}: Efficient Long-Context {LLM} Serving with Head-Aware {KV} Reuse and {SegPagedAttention}},
  author={Liu, Yang and Luo, ZhaoKai and Jin, HuaYi and Wang, ZhiYong and He, RuoZhou and Wang, BoYu and Chen, Guanjie and Hu, Junhao},
  year={2026},
  eprint={2606.06256},
  archivePrefix={arXiv},
  url={https://arxiv.org/abs/2606.06256}
}

@article{deepseekv3,
  title={{DeepSeek-V3} Technical Report},
  author={{DeepSeek-AI}},
  journal={arXiv preprint arXiv:2412.19437},
  year={2024}
}

\end{document}